\pdfoutput=1
\documentclass[12pt]{l4dc2021}

\usepackage[export]{adjustbox}
\usepackage{mathtools}
\usepackage{natbib}

\usepackage{multicol}
\usepackage{bm}
\usepackage{amsmath}
\usepackage{amsfonts}
\usepackage{graphicx}

\usepackage{cleveref}

\usepackage{booktabs}
\usepackage{multirow}
\usepackage{tikz}
\usepackage{adjustbox}
\usepackage{floatflt}
\usepackage{adjustbox}
\usepackage{wrapfig}
\usepackage{algorithm}
\usepackage{algorithmic}

\newcommand{\bb}{\mathbf}

\begin{document}
\title{
L4KDE:
Learning for KinoDynamic Tree Expansion
}

\author{%
 \Name{Tin Lai} \Email{tin.lai@sydney.edu.au}\\
 \addr School of Computer Science, The University of Sydney
 \AND
 \Name{Weiming Zhi} \Email{weiming.zhi@sydney.edu.au}\\
 \addr School of Computer Science, The University of Sydney
 \AND
 \Name{Tucker Hermans} \Email{tucker.hermans@utah.edu}\\
 \addr School of Computing, University of Utah and NVIDIA, USA
  \AND
 \Name{Fabio Ramos} \Email{fabio.ramos@sydney.edu.au}\\
 \addr School of Computer Science, The University of Sydney and NVIDIA, USA
}

\maketitle

\begin{abstract}
We present the \textbf{L}earning for \textbf{K}ino\textbf{D}ynamic Tree \textbf{E}xpansion (L4KDE) method for kinodynamic planning. Tree-based planning approaches, such as rapidly exploring random tree (RRT), are the dominant approach to finding globally optimal plans in continuous state-space motion planning. Central to these approaches is tree-expansion, the procedure in which new nodes are added into an ever-expanding tree. We study the kinodynamic variants of tree-based planning, where we have known system dynamics and kinematic constraints. In the interest of quickly selecting nodes to connect newly sampled coordinates, existing methods typically cannot optimise to find nodes that have low cost to transition to sampled coordinates. Instead, they use metrics like Euclidean distance between coordinates as a heuristic for selecting candidate nodes to connect to the search tree. We propose L4KDE to address this issue. L4KDE uses a neural network to predict transition costs between queried states, which can be efficiently computed in batch, providing much higher quality estimates of transition cost compared to commonly used heuristics while maintaining almost-surely asymptotic optimality guarantee. We empirically demonstrate the significant performance improvement provided by L4KDE on a variety of challenging system dynamics, with the ability to generalise across different instances of the same model class, and in conjunction with a suite of modern tree-based motion planners.

\end{abstract}

\newcommand*\state{\mathbf{q}}
\newcommand*\stateOrigin{{\state_\mathrm{o}}}
\newcommand*\State{\mathcal{Q}}
\newcommand*\OracleCost{\mathcal{L}}
\newcommand*\LearnedCost{{\mathcal{L}_\theta}}
\newcommand*\Transform{\mathcal{T}_\stateOrigin}

\section{Introduction}
This paper focuses on asymptotically optimal, kinodynamic motion planning, where one aims to find a globally-optimal plan between start and goal coordinates while remaining both (1) collision-free, and (2) compliant to the kinodynamic constraints of the robot or control system. When the search-space is continuous, kinodynamic variants of tree-based motion planning algorithms are often the main workhorse for tackling these problems. Existing tree-based methods often consider both collision and kinodynamic constraints jointly during the motion planning procedure. However, if one disentangles the two constraints, one notices that collision constraints are \emph{environment-specific}, while kinodynamic constraints are inherited from the robot hardware, and do not vary over different planning tasks.

Our motivation stems from the critical need for efficient and globally optimal motion planning in the context of kinodynamic systems.
Kinodynamic motion planning is a fundamental problem in robotics, where robots must navigate through complex environments while adhering to both collision avoidance and kinematic constraints.
Achieving this balance is essential for real-world applications like autonomous vehicles, industrial robots, and humanoid robots.
In particular, unlike the assumption of holonomic robots, practical, real-world robots are often subjected to some form of kinodynamic constraints.
The choice of nodes directly impacts the efficiency and quality of the resulting motion plan, where existing methods typically rely on simplistic heuristics, like Euclidean distance, to guide node selection.
By providing a more accurate estimation of state-transition cost, tree expansion can often benefit from successful and realistic distance estimation during state forward propagation, thereby expanding the capabilities of kinodynamic planners.

With this insight, we present the \textbf{L}earning for \textbf{K}ino\textbf{D}ynamic Tree \textbf{E}xpansion (L4KDE) method within tree-based motion planners. L4KDE utilises a neural network to predict the cost of transitioning between states under kinodynamic constraints, allowing for efficient retrieval of the cost when planning online. By leveraging the high representation capacity of neural networks, L4KDE is exceptionally novel in that it can also condition directly on parameters of the dynamics model (For example, velocity limits). By employing the cost as a node selection heuristic in tree expansion of the underlying planning algorithm, we obtain significant improvements in planning performance over the commonly used Euclidean heuristic. We show that L4KDE consistently improves performance when used within a wide range of tree-based motion planning algorithms across a variety of tasks. Importantly we show that the improvement provided by L4KDE holds for state-of-the-art, tree-based, asymptotically optimal kinodynamic motion planners~\citep{hauser2016asymptotically,li2016asymptotically}.

\section{Related Work}

Kinodynamic planning refers to the class of robot systems that must obey kinematic and differential constraints~\citep{donald1993kinodynamic}.
For example velocity, acceleration or force bounds in system such as nonholonomic vehicles~\citep{likhachev2009planning} and locomotion systems \citep{kuindersma2014efficiently}.
\emph{Sampling-based motion planners} (SBPs) are a class of algorithms that avoid explicit representation of state space while connecting feasible states to form solution trajectories.
Compared to solutions given by trajectory optimisation techniques which are prone to local minima~\citep{betts1998survey}, SBPs provides \emph{probabilistic completeness}~\citep{verscheure2009time} and \emph{almost-surely asymptotic optimality} guarantee~\citep{elbanhawi2014_SampRobo}.

Tree-based methods such as RRT~\citep{lavalle1998rapidly} are the predominant methods in SBPs.
RRTs utilise random sampling to create Voronoi bias~\citep{lindemann2004incrementally}, which helps to explore the space rapidly~\citep{lai2018_BalaGlob}.
Compared to roadmap-based methods such as PRM~\citep{kavraki1996_ProbRoad}, tree-based methods do not require a steering function, which requires solving the two-point boundary value problem.
Techniques that focus on sampling-based planning can either focus on the algorithmic parts of graph construction~\citep{lai2022ltr} or on the sampling aspect of the planner~\citep{ichter2018_LearSamp}.
Experience-based approaches can also improve planning efficiency,
for example, the
Lightning~\citep{berenson2012_RoboPath} and Thunder~\citep{coleman2015experience} framework uses database structure to store past trajectories.
However, most of these approaches only focus on kinematic planning and are not directly applicable to kinodynamic planning. Additionally, L4KDE is able to generalise over entire classes of dynamics models, where each may have different constraint limits or parameter values, which are conditioned on.

For kinodynamic motion planning, we can form a suitable steering function with LQR in systems dynamics that can be linearised by quadratic functions~\citep{perez2012lqr}.
Some approaches use techniques to discover narrow passages~\citep{lai2020_BayeLoca,lai2021_AdapExpl}, uses space projection~\citep{orthey2019_RapiQuot} or heuristic measures of obstacles boundary~\citep{
lee2012_SRRRSele} to improve sampling.
However, most approach does not works with kinodynamic constraints.
There has been also numerous learning approaches, such as leveraging experience or environment \citep{lai2021_PlanLear} to utilise learned neural networks for sampling.
However, this is often not directly applicable for the general class of
kinodynamic problems when more complex
differential constraints
are involved.

Learning-based approach have been one of the main focus in the sampling procedure of SBPs, for example, by learning sampling  distribution~\citep{ichter2018_LearSamp} from previous experience or utilising obstacle information to form an informed distribution~\citep{lai2021parallelised}.
However, there is a lack of study in the node selection procedure as most existing works uses Euclidean distance to guide tree expansion.
Moreover, there are a few works dedicated to asymptotically optimal, kinodynamic planning~\citep{kunz2014probabilistically,li2016asymptotically,hauser2016asymptotically} but most continue to use the ill-formed Euclidean distance metric to expand nodes in their search trees. In this work, we attempt to tackle this issue via formalising the learning-based tree-expansion procedure under the kinodynamic planning framework.

\section{Kinodynamic Tree-based Motion Planning}

\subsection{Problem Setup}
We shall first outline the kinodynamic motion planning problem. Let the state space be denoted as $\mathcal{X}$, the set of controls within the control constraints as $\mathcal{U}$, the set of states where kinematic constraints are satisfied as $\mathcal{K}$, and the set of collision-free states as $\mathcal{O}$.
\begin{definition}
A feasible kinodynamic planning problem asks, given an initial state $\bb{x}_{0}$, goal state $\bb{x}_{goal}$, system dynamics given by $f$, to produce a trajectory $\bb{x}(t):[0, t_{max}]\rightarrow \mathcal{X}$ and a sequence of controls $\bb{u}(t):[0,t_{max}]\rightarrow\mathcal{U}$, such that:

\begin{align}
    \bb{x}(0) &= \bb{x}_{0}, && \text{(initial state)}\label{constr1}\\
    \bb{x}(t_{max}) &= \bb{x}_{goal}, && \text{(goal state)}\label{constr2}\\
    \bb{x}(t)& \in \mathcal{K} \subseteq \mathcal{X},  \forall t\in[0,t_{max}], && \text{(kinematic constraints)}\label{constr3}\\
    \bb{x}(t)& \in \mathcal{O} \subseteq \mathcal{X},  \forall t\in[0,t_{max}], && \text{(obstacle constraints)}\label{constr4}\\
    \bb{x}'(t)& = f(\bb{x}(t), \bb{u}(t)), \forall t\in[0,t_{max}], && \text{(dynamics model)}\label{constr5}
\end{align}
\end{definition}

The \emph{optimal kinodynamic planning problem}, additionally asks to find a trajectory which minimises an objective cost, subject to the constraints in \cref{constr1,constr2,constr3,constr4,constr5}. This objective is given by:
\begin{equation}
    C(\bb{x}_{0}, \bb{x})=\int_{0}^{t_{max}}L(\bb{x}(t),\bb{u}(t))\mathrm{d}t+\Phi(\bb{x}(t_{max})),\label{obj_cost}
\end{equation}
where $L$ is the incremental cost, and $\Phi$ is the terminal cost.

\subsection{Tree-Expansion}
Tree-based motion planners like RRT$^{*}$~\citep{KaramanIJRR2011} iteratively connect new coordinates into a space-filling tree, by randomly sampling and checking coordinates, before expanding the tree to connect with valid samples. This typically requires a \emph{steering function} which finds the optimal path between two coordinates under the presence of no obstacle---in the absence of kinodynamic constraints, this is simply finding a straight line. Although a steering function may be available for a restrictive class of kinodynamically-constrained system, we cannot generally find the optimal path without resorting to solving an optimal kinodynamic planning problem between a tree node and the sampled point.

To address this challenge, kinodynamic variants of tree-based methods have been introduced. These methods do not seek to connect sampled coordinates deemed valid, but simplify the optimisation and attempt to forward propagate, from the nearest existing tree node, in a kinodynamically feasible manner to a viable coordinate ``closest enough'' to the sample, where the viable coordinate to the tree in lieu of the sample.

The algorithmic outline of the tree expansion in shown in \cref{treeEx}. In the tree-building process, provided a randomly sampled coordinate, we:
(i) find the nearest node, $\bb{x}_{active}$, in the existing tree $\mathcal{T}$, via $\mathtt{Nearest}(\cdot)$;
(ii) propagate trajectories by holding a valid control from $\bb{x}_{active}$, for a randomly sampled amount of time, to find a candidate ``close'' coordinate $\bb{x}_{candidate}$. This is typically performed by either randomly selecting a control, or randomly sampling and holding multiple controls and selecting the sampled control where the final state is the closest to $\bb{x}_{sample}$;
(iii) validate whether $\bb{x}_{candidate}$ can be feasibly connected to $\bb{x}_{active}$, by checking if the connection satisfies collision constraints, and if the cost to reach the new node is below the current best cost $c^{*}$;
(iv) connect $\bb{x}_{candidate}$ as a new node to $\mathcal{T}$.
The tree expansion process is run iteratively, as we sample more coordinates as $\bb{x}_{sample}$.

\begin{algorithm}[t]
    \caption{Kinodynamic Tree-Expansion}\label{treeEx}

    \begin{algorithmic}
    \REQUIRE $\mathcal{T}$, $\bb{x}_{sample}$, $\mathtt{HoldControl}(\cdot,\cdot)$, $\mathtt{Nearest}(\cdot)$, $\mathcal{U}$, $C(\cdot, \cdot)$, $c^{*}$, $\mathtt{CollisionCheck}(\cdot)$, $\mathtt{ExpandTree}(\cdot,\cdot,\cdot)$
    \STATE $\bb{x}_{active}\leftarrow \mathtt{Nearest}(\bb{x}_{sample})$
    \STATE $\bb{x}_{candidate}\leftarrow \mathtt{HoldControl}(\bb{x}_{active},\bb{u},t)$
    \IF{$(\mathtt{CollisionCheck}(\bb{x}_{candidate}))\And (C(\bb{x}_{0}, \bb{x}_{candidate})<c^{*})$}
    \STATE $\mathcal{T}\leftarrow \mathtt{ConnectNewNode}(\mathcal{T},\bb{x}_{active},\bb{x}_{candidate})$
    \ENDIF
    \end{algorithmic}
\end{algorithm}

\section{L4KDE}
In this section, we introduce our method of kinodynamic tree expansion, \textbf{L}earning for \textbf{K}ino\textbf{D}ynamic Tree \textbf{E}xpansion (L4KDE). In the following sections, we
    (i) outline the flaws of the existing active node selection heuristic in tree-based planning in the kinodynamic setup, where Euclidean distance or absolute angular difference is used as a proxy for transition cost~\citep{hauser2016asymptotically};
    (ii) describe a learning-based approach to quickly predict transition costs for the selection of active nodes;
    (iii) elaborate on training a neural network to predict state transition costs.

\begin{figure}[bt]
    \centering
        \subfigure[Existing tree-based kinodynamic methods select the active node with Euclidean distance.
        The two coordinates are close in Euclidean space, but moving from the left to the destination on the right assuming Dubins car dynamics may be fairly costly to reach (via red trajectory).]
        {%
            \label{dubins_example}
            \includegraphics[width=.46\linewidth]{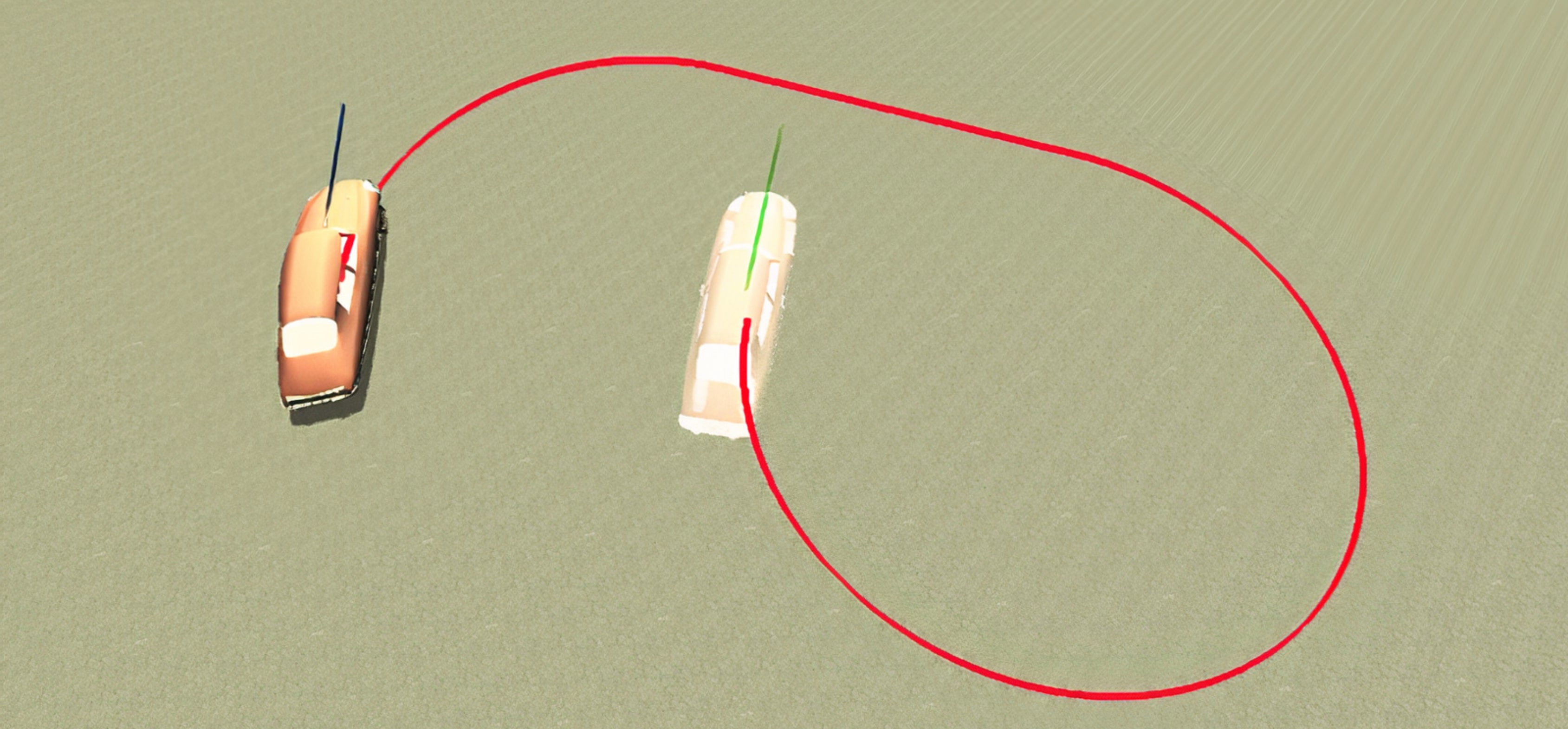}%
        }\hfill
    \subfigure[Transition costs between the origin state ($\theta_1=\pi, \theta_2=\pi$) and other coordinates in state-space, for the \emph{double pendulum} (with 4D state-space).
    Axes shows the the two joint angles $\theta_1, \theta_2$ and the angular velocity of the first joint $\omega_1$;
    For visualisation purposes, the corresponding $\omega_2$ is omitted.]
        {
            \label{fig:double-pendulum-plot}
            \includegraphics[width=.46\linewidth]{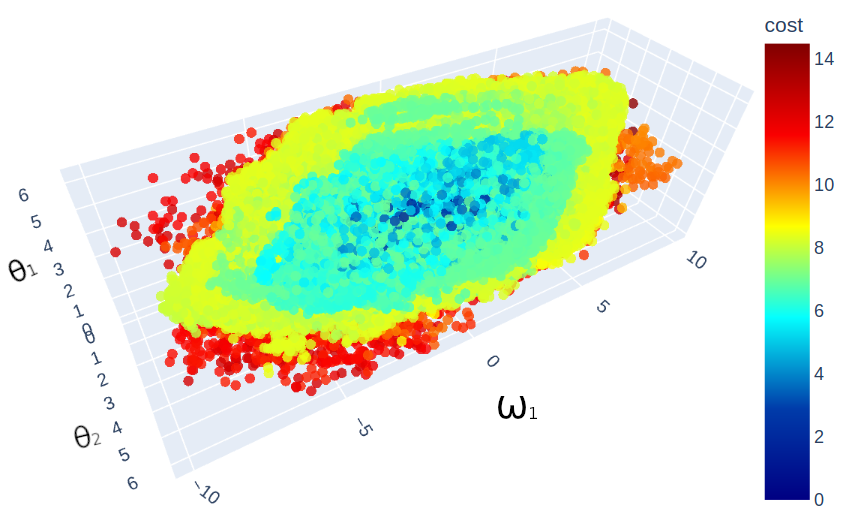}%
        }
    \caption{%
        Illustrations of (a) the problem with an ill-informed metric and (b) visualisation of learning the differential system dynamics with the proposed L4KDE.
        Existing tree-based kinodynamic methods select the active node with Eulidean distance. The two coordinates are close in Euclidean space, but moving from the current coordinates on the left to the destination on the right assuming Dubins car dynamics may be fairly costly to reach (via red trajectory).
    }
\end{figure}

\subsection{Euclidean-based Heuristics: Fast but Flawed}
The heuristic $\mathtt{Nearest}(\cdot)$ aims to find the node in the existing tree which is the nearest to the sampled point, under kinematic constraints. From \cref{treeEx}, we can see that this is decoupled from the collision-checking. When calling $\mathtt{Nearest}(\cdot)$ during tree-expansion, existing kinodynamic tree-based planning methods simply select the closest node coordinates from the tree, $\mathcal{T}$, using Euclidean distance,
\begin{align}
\bb{x}_{active}&=\mathtt{Nearest}(\bb{x}_{sample}) \\ &=\arg\min_{\bb{x}\in\mathcal{T}}\lvert\lvert \bb{x}-\bb{x}_{sample} \lvert\lvert_{2}. \label{minl2}
\end{align}
In the absence of kinodynamic constraints, choosing the active node via minimising the Euclidean distance is guaranteed to provide the lowest cost path to the sampled coordinate. However, under kinodynamic constraints, this is often not the case, and a high cost may be required to transition to coordinates close in Euclidean space. We illustrate in \cref{dubins_example} an example of coordinates which are close in Euclidean distance, but are fairly costly to reach under the Dubins car dynamics. In later sections, we shall empirically demonstrate that in many common kinodynamic planning tasks the nearest node according to Euclidean distance is frequently costly to transition to.

\subsection{Transition Cost Prediction for Tree Expansion}

Would it then be possible to find the nearest point, being the node on the existing tree $\mathcal{T}$ which can reach $\bb{x}_{sample}$ while incurring the lowest cost? Computing the (lowest) cost to kinodynamically-feasibly transition between two coordinates in the absence of obstacles requires solving the optimal cost kinodynamic planning problem between nodes in the tree and the sampled coordinate. Finding the nearest node based on transition cost requires us to solve the constrained optimisation problem:
\begin{align}
    \bb{x}_{active} &=\!\arg\min_{\bb{x}\in\mathcal{T}}\mathtt{\!TransitionCost\!}(\bb{x}\!, \!\bb{x}_{sample})\!, \label{nearest_obj}\\
    \mathtt{Transition}&\mathtt{Cost}(\bb{x}, \bb{x}_{sample}) = \min_{\bb{u}\in\mathcal{U}} C(\bb{x}, \bb{x}_{sample})\label{nearest_subobj}\\
   \text{s.t. } &\bb{x}(0) = \bb{x},\; \bb{x}(t_{max}) = \bb{x}_{sample},\label{nearest_con1}\\
   &\bb{x}(t) \in \mathcal{K} \subseteq \mathcal{X}, \; \\
   &\bb{x}'(t) = f(\bb{x}(t), \bb{u}(t)),  \forall t\in[0,t_{max}]. \label{nearest_con2}
\end{align}

In general, finding the cost of a feasible trajectory between nodes in the existing tree and the newly sampled node cannot be efficiently solved online. We propose to instead approximate the obstacle-free transition cost function in \cref{nearest_subobj}, via offline learning with a function approximator, and minimise the approximate cost online to select the active node. Importantly, as the cost function factors in the kinodynamic constraints, but not the environment-specific collision-avoidance constraints, the trained model is specific to the system dynamics and does not need to be tuned when planning in different environments. We use a neural network, denoted as $g_{\bm{\theta}}$, with parameters $\bm{\theta}$ as the function approximator. Finding the nearest node by the approximate transition cost is then given as the unconstrained:
\begin{align}
\bb{x}_{active}=\arg\min_{\bb{x}\in\mathcal{T}} g_{\bm{\theta}}(\bb{x},\bb{x}_{sample}). \\ g_{\bm{\theta}} \approx \mathtt{TransitionCost}((\bb{x}, \bb{x}_{sample}).  \label{min_nn}
\end{align}
This can be solved very efficiently, as it only requires one batched query of $g_{\bm{\theta}}$, then selecting the $\arg\min$ from the outputs. The batched querying of neural networks can be performed efficiently in parallel on a GPU. Empirically, we found the run-time of finding $\bb{x}_{active}$ via \cref{min_nn} to be roughly twice as long as finding $\bb{x}_{active}$ via \cref{minl2}, as is done in existing planning methods.
It is important to note that our approach continues to maintain almost-surely asymptotic optimality guarantee for our various optimal planners as our procedure does not alter their subroutine optimality convergence.
We had omitted the discussion in this work for brevity.

\begin{table*}[t]
\caption{Time-to-solution and success percentage for the Dubins Car turn-rate experiments.\label{table:turn-rate}}
\centering
\resizebox{0.95\linewidth}{!}{%
\begin{tabular}{@{}cccccccccc@{}}
\toprule
\multirow{3}{*}{} & \multirow{3}{*}{Motion Planner} & \multicolumn{8}{c}{Dubins Car Maximum turning rate ($\mu \pm \sigma$) } \\ \cmidrule(l){3-10}
 &  & \multicolumn{2}{c}{$\Omega = 0.5$} & \multicolumn{2}{c}{$\Omega = 1.38$} & \multicolumn{2}{c}{$\Omega = 2.26$} & \multicolumn{2}{c}{$\Omega = \pi$} \\ \cmidrule(l){3-10}
 &  & Standard & with L4KDE & Standard & with L4KDE & Standard & with L4KDE & Standard & with L4KDE \\ \midrule
\multirow{6}{*}{\rotatebox[origin=c]{90}{Time-to-sol.}} & \multicolumn{1}{c|}{RRT} & 10.84 ± 7.55 & 7.98 ± 6.81 & 4.07 ± 4.02 & 2.38 ± 1.75 & 2.66 ± 3.34 & 0.96 ± 0.93 & 1.39 ± 0.95 & 2.05 ± 1.79 \\
 & \multicolumn{1}{c|}{AO-RRT} & 17.99 ± 3.89 & 7.45 ± 7.86 & 8.75 ± 9.27 & 2.74 ± 1.82 & 4.29 ± 5.17 & \textbf{0.64 ± 0.40} & 3.58 ± 3.01 & \textbf{0.71 ± 0.60} \\
 & \multicolumn{1}{c|}{Anytime-RRT} & 15.31 ± 4.65 & \textbf{6.95 ± 4.58} & 5.84 ± 4.52 & 2.35 ± 1.85 & 3.63 ± 2.68 & 1.33 ± 1.07 & 3.93 ± 3.05 & 2.26 ± 2.04 \\
 & \multicolumn{1}{c|}{SST} & - & 9.05 ± 5.81 & 6.40 ± 6.32 & \textbf{1.31 ± 0.97} & 5.52 ± 6.54 & 1.59 ± 1.34 & 3.86 ± 4.03 & 1.59 ± 1.69 \\
 & \multicolumn{1}{c|}{SST*} & - & 20.50 ± 5.64 & 8.29 ± 6.43 & 3.33 ± 2.57 & 4.91 ± 4.52 & 1.37 ± 1.58 & 3.76 ± 4.50 & 2.10 ± 1.76 \\
 & \multicolumn{1}{c|}{AO-EST} & - & N/A & 20.70 ± 5.41 & N/A & 17.23 ± 5.68 & N/A & 11.84 ± 5.95 & N/A \\ \midrule
\multirow{6}{*}{\rotatebox[origin=c]{90}{Success pct.}} & \multicolumn{1}{c|}{RRT} & 65.0\% & 70.0\% & 100.0\% & 100.0\% & 100.0\% & 100.0\% & 85.0\% & 100.0\% \\
 & \multicolumn{1}{c|}{AO-RRT} & 10.0\% & 85.0\% & 95.0\% & 100.0\% & 95.0\% & 100.0\% & 95.0\% & 100.0\% \\
 & \multicolumn{1}{c|}{Anytime-RRT} & 40.0\% & 50.0\% & 80.0\% & 100\% & 90.0\% & 100\% & 95.0\% & 100\% \\
 & \multicolumn{1}{c|}{SST} & 0.0\% & 85.0\% & 95.0\% & 100\% & 100.0\% & 100.0\% & 95.0\% & 100.0\% \\
 & \multicolumn{1}{c|}{SST*} & 0.0\% & 45.0\% & 85.0\% & 100.0\% & 100.0\% & 100.0\% & 100.0\% & 100.0\% \\
 & \multicolumn{1}{c|}{AO-EST} & 0.0\% & N/A & 75.0\% & N/A & 90.0\% & N/A & 95.0\% & N/A \\ \bottomrule
\end{tabular}
}
\end{table*}

\begin{figure}[tb]
    \centering%
    \subfigure[Dubins car. Red circle denotes goal.]{
        \label{fig:dubins-car}
        \includegraphics[width=0.495\linewidth]{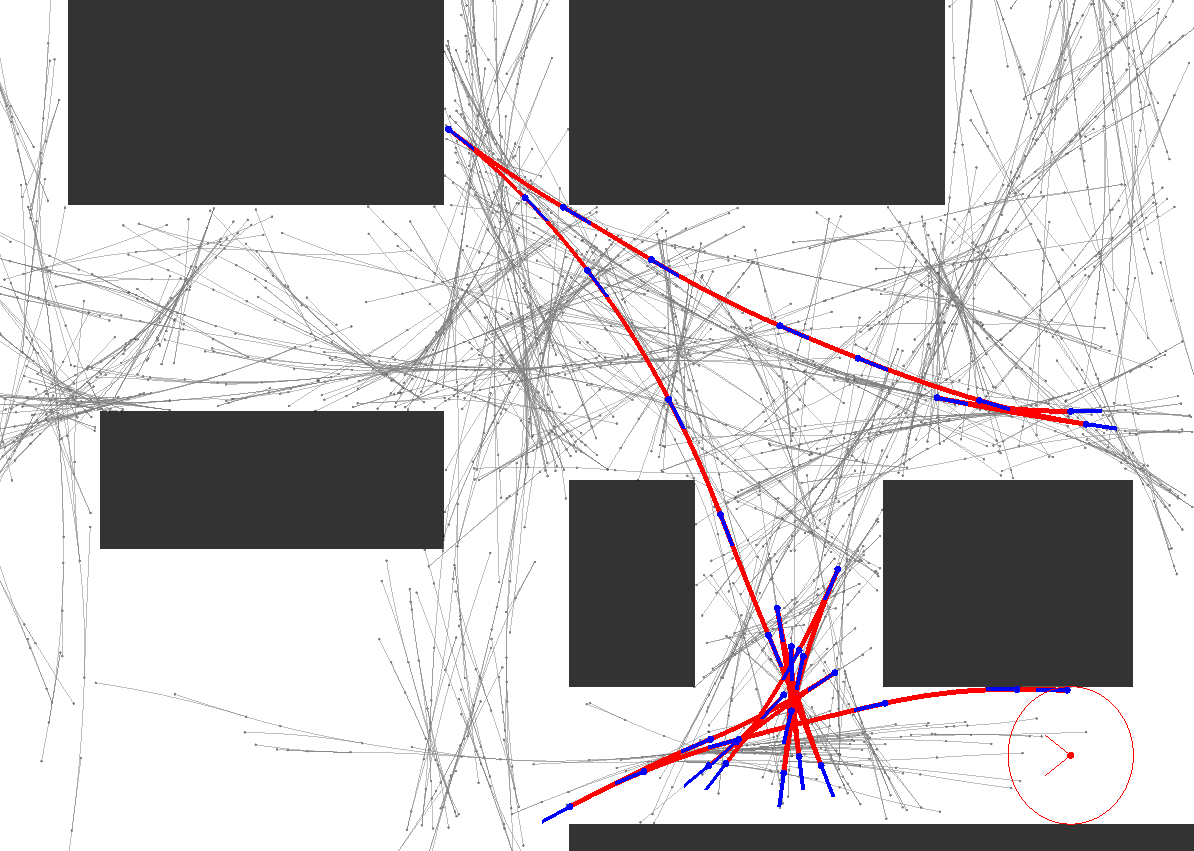}
    }\hfill%
    \subfigure[Gooseneck trailer. Red circle denotes goal.]{
        \label{fig:dubins-trailer}
        \includegraphics[width=0.36\linewidth,trim={0 0 0 3.5cm},clip]{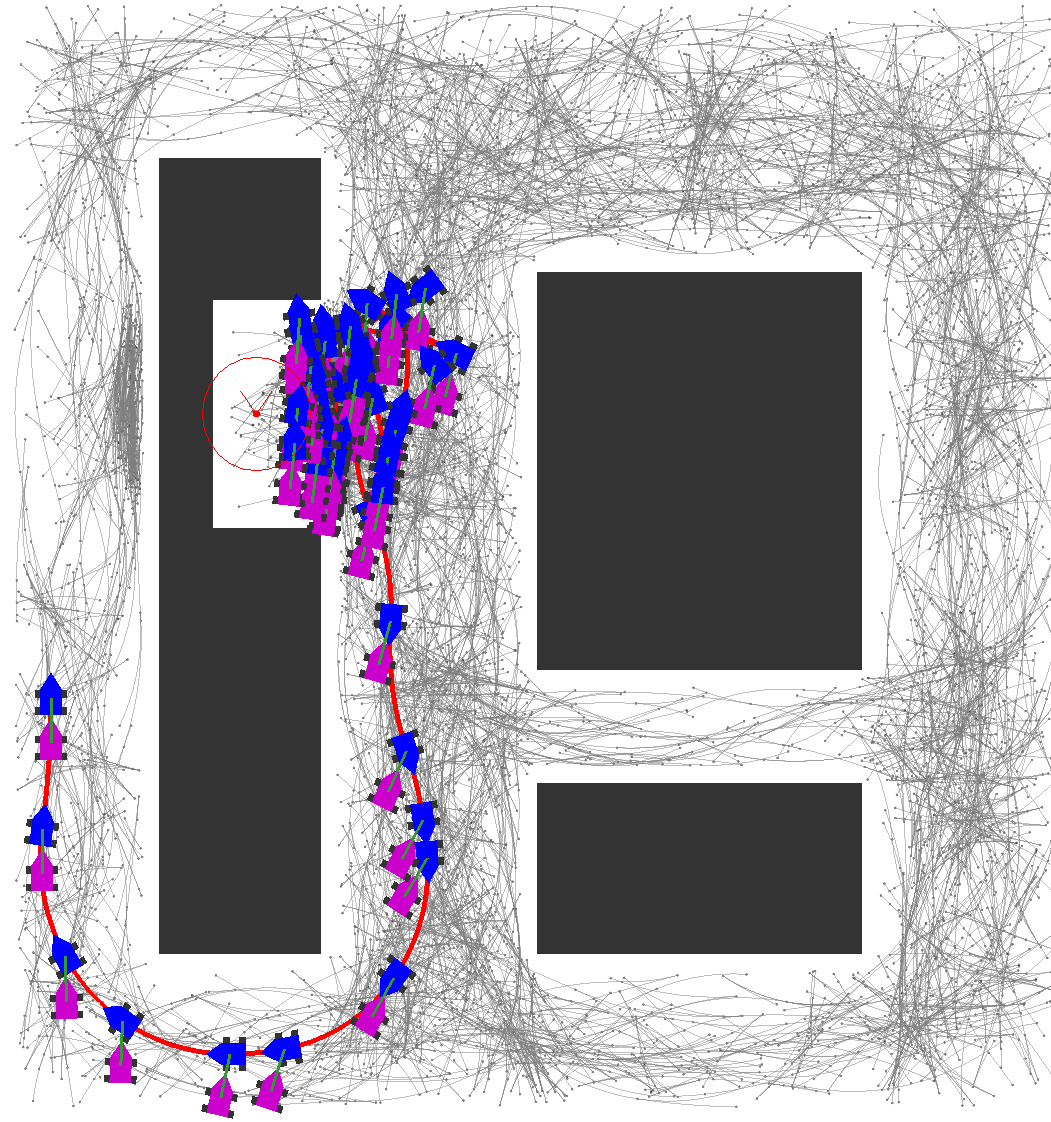}
    }
\hfill\subfigure[DP during planning. Semi-transparent lines denotes tree edges (currently reachable set).]{
        \label{fig:2JP-start}
        \includegraphics[width=0.46\linewidth]{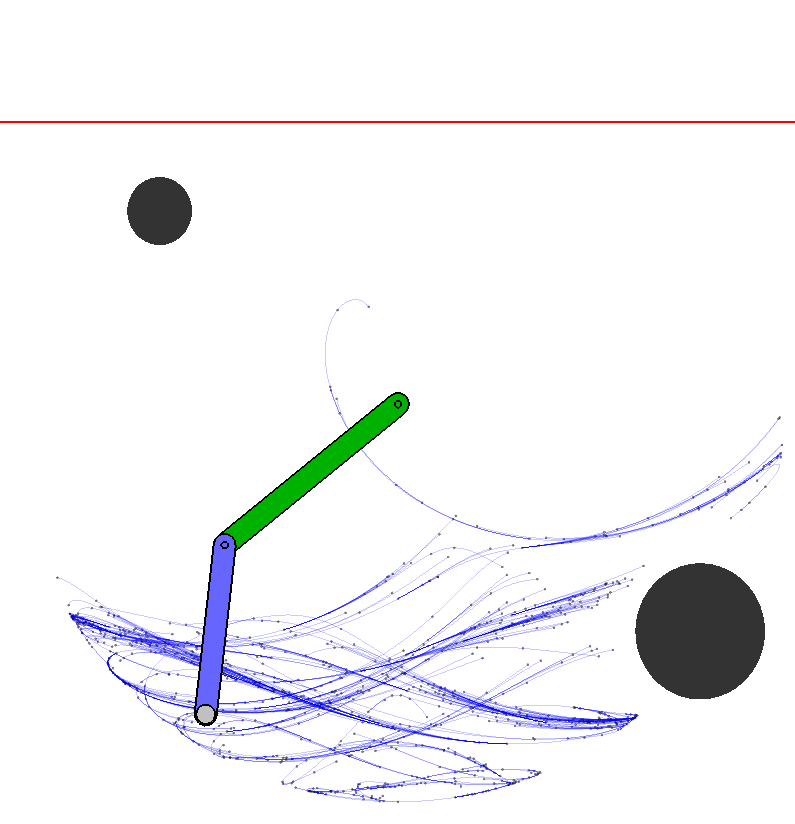}
    }\hspace{0.2cm}%
    \subfigure[DP with a successful trajectory (reaching the horizontal red line, which denotes vertical altitude goal).]{
        \label{fig:double-joint}
        \includegraphics[width=0.46\linewidth]{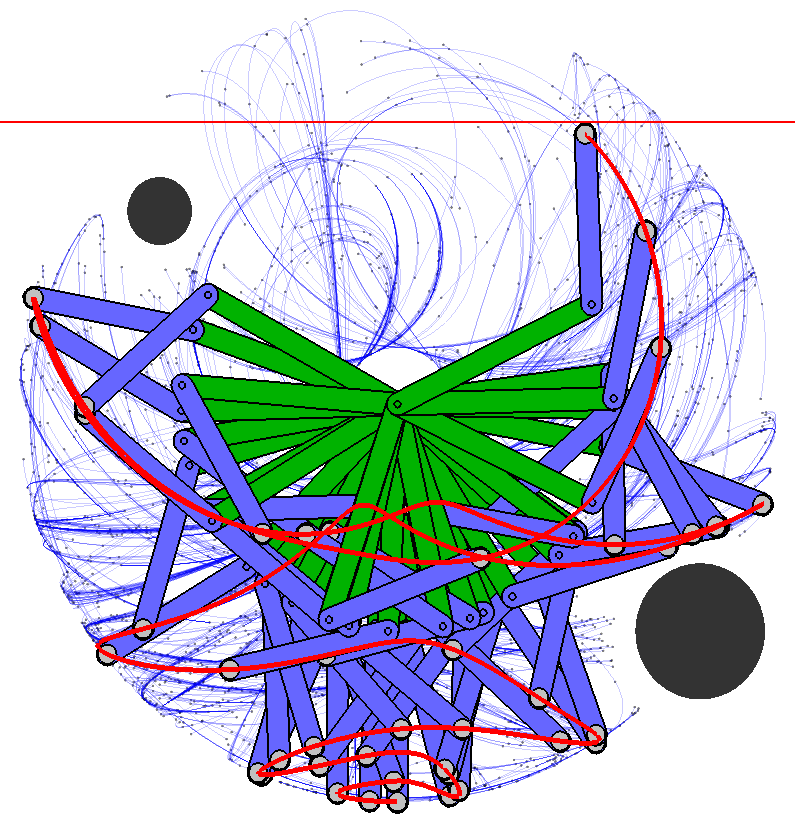}
    }\hfill

    \caption{Planning environments. Solution trajectories are shown in red; tree in grey ((a) and (b)) and blue ((c) and (d)); obstacles in black. Subfigure (c) and (d) illustrates the Double Pendulum (DP) problem with obstacles.}
\end{figure}

\subsection{Training the Function Approximator}

As the kinematic constraints are not tied to a specific environment, we can train the function approximator offline $g_{\theta}: \mathcal{S}\times\mathcal{S}\rightarrow \mathbb{R}$ to approximate the transition cost. We collect a dataset with $n$ data examples $\mathcal{D}=\{(\bb{x}^{i}_{start},\bb{x}^{i}_{goal}), c^{i}\}_{i=1}^{n}$, where $c$ is the cost to the goal, $\bb{x}_{goal}$, from the start $\bb{x}_{start}$. This dataset can come either from offline simulation or by observing planning runs online in a life-long learning manner.
As stated above we use a neural network as the function approximator $g_{\bm{\theta}}$. Neural networks are universal function approximators \citep{sonoda2017neural}, and have been widely used on learning problems involving large and high-dimensional datasets. We train the neural network via mean squared error loss:
\begin{equation}
    \mathcal{L}(\bm{\theta})=\lvert\lvert c-g_{\bm{\theta}}(\bb{x}_{start}\oplus\bb{x}_{goal}) \lvert\lvert_{2}^{2},
\end{equation}
where $\bb{x}_{start}\oplus\bb{x}_{goal}$ denotes the vector concatenation of the start and goal coordinates. The model $g_{\bm{\theta}}$ can then be trained via gradient descent optimisers.

Typically the dataset $\mathcal{D}$ is collected from a single system, and the learned transition cost predictor will be specific to dynamics model and kinematics constraints of that system. Importantly, we can learn a transition cost predictor to a class of dynamic models with varying kinematic constraints by conditioning on the limit values of the constraints. For example, we can learn a cost predictor model for Dubins car models with various turn rate limits, which we denote as $\Omega$. We then augment the input space of the neural network by concatenating the turn rate, i.e. $g_{\bm{\theta}}(\bb{x}_{start}\oplus\bb{x}_{goal}\oplus\Omega)$.
The results are shown in~\cref{table:turn-rate}.

\begin{figure}[h]
    \centering%
    \includegraphics[width=\linewidth]{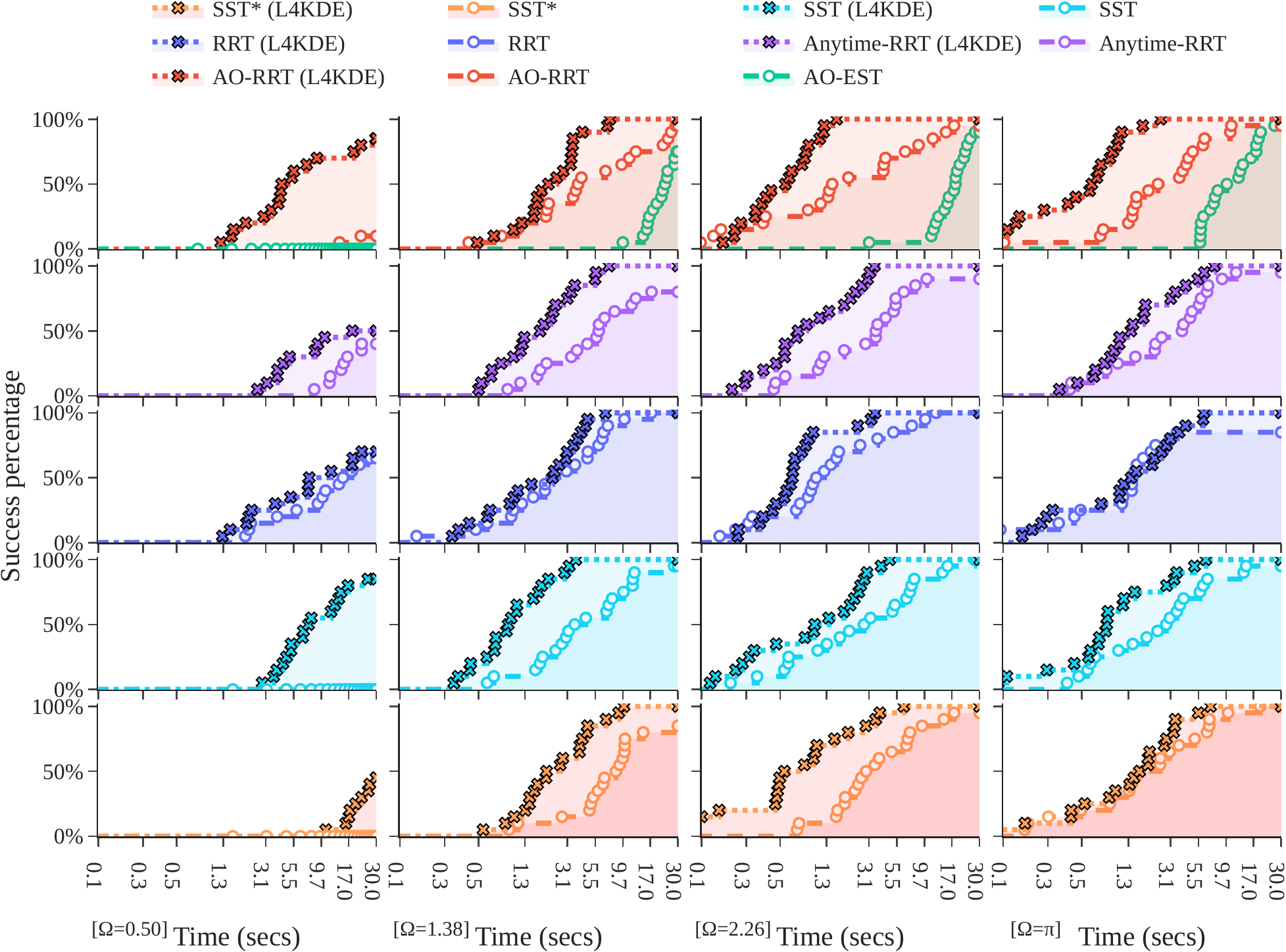}
    \caption{%
        Success percentage for various max turn-rate $\Omega$ in the Dubins car environment.
        \label{fig:result-dubins}
    }
\end{figure}

\begin{figure}[h]
    \centering%
    \includegraphics[width=\linewidth]{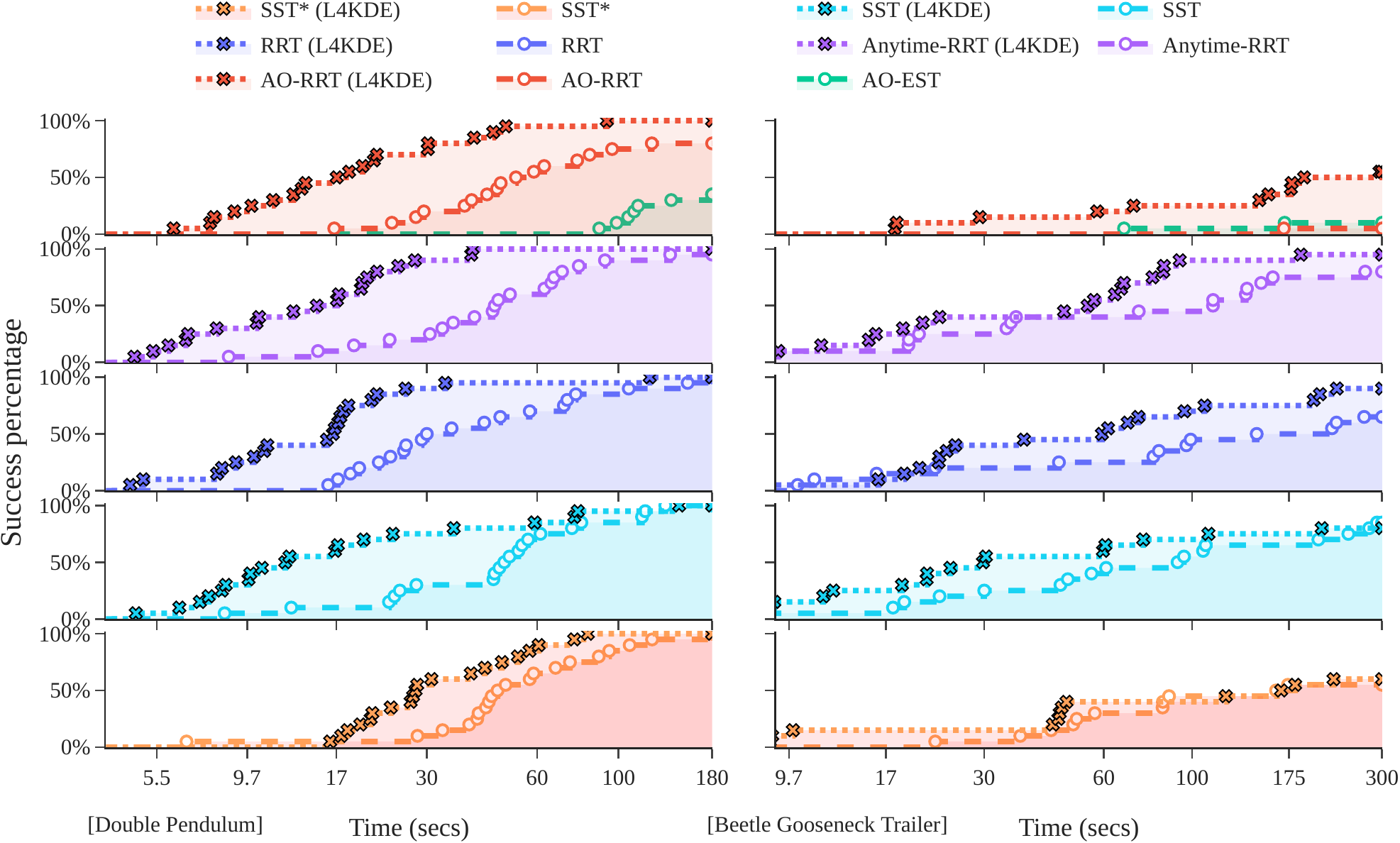}
    \caption{%
        Success percentage in the Double Pendulum and Beetle Gooseneck Trailer environment over 20 runs.
        \label{fig:result-other-env}
    }
\end{figure}

\begin{table*}[ht]
\centering
\caption{Time-to-solution ($\mu \pm \sigma$) and success percentage for the double pendulum and beetle gooseneck trailer
}\label{table-hard-problems}

\begin{tabular}{cccc}
\toprule
 & \multirow{2}{*}{Motion Planner} & \multicolumn{2}{c}{Double Pendulum} \\ \cmidrule{3-4}
 &  & Standard & with L4KDC \\ \midrule
\multirow{6}{*}{\rotatebox[origin=c]{90}{Time-to-sol.}} & \multicolumn{1}{c|}{RRT} & 53.58 ± 44.76 & 20.58 ± 24.21 \\
 & \multicolumn{1}{c|}{AO-RRT} & 54.23 ± 27.69 & 23.99 ± 20.06 \\
 & \multicolumn{1}{c|}{Anytime-RRT} & 50.94 ± 29.90 & \textbf{16.76 ± 10.21} \\
 & \multicolumn{1}{c|}{SST} & 55.41 ± 33.71 & 28.52 ± 34.72 \\
 & \multicolumn{1}{c|}{SST*} & 63.04 ± 37.16 & 36.97 ± 19.31 \\
 & \multicolumn{1}{c|}{AO-EST} & 119.39 ± 28.57 & N/A \\ \midrule
\multirow{6}{*}{\rotatebox[origin=c]{90}{Success pct.}} & \multicolumn{1}{c|}{RRT} & 100.0\% & 100.0\% \\
 & \multicolumn{1}{c|}{AO-RRT} & 80.0\% & 100.0\% \\
 & \multicolumn{1}{c|}{Anytime-RRT} & 95.0\% & 100.0\% \\
 & \multicolumn{1}{c|}{SST} & 100.0\% & 100.0\% \\
 & \multicolumn{1}{c|}{SST*} & 100.0\% & 100.0\% \\
 & \multicolumn{1}{c|}{AO-EST} & 35.0\% & N/A \\ \bottomrule
\end{tabular}

\vspace{1em}

\begin{tabular}{cccc}
\toprule
 & \multirow{2}{*}{Motion Planner} & \multicolumn{2}{c}{Beetle Gooseneck Trailer} \\ \cmidrule{3-4}
 &  & Standard & with L4KDC \\ \midrule
\multirow{6}{*}{\rotatebox[origin=c]{90}{Time-to-sol.}} & \multicolumn{1}{c|}{RRT} &  171.76 ± 116.88 & 95.04 ± 94.92  \\
 & \multicolumn{1}{c|}{AO-RRT} &  293.50 ± 28.35 & 201.93 ± 108.89  \\
 & \multicolumn{1}{c|}{Anytime-RRT} &  126.43 ± 108.44 & \textbf{65.31 ± 68.08}  \\
 & \multicolumn{1}{c|}{SST} &  131.50 ± 110.57 & 95.17 ± 112.39  \\
 & \multicolumn{1}{c|}{SST*} &  177.74 ± 115.96 & 167.84 ± 121.38  \\
 & \multicolumn{1}{c|}{AO-EST} & 281.91 ± 56.69 & N/A  \\ \midrule
\multirow{6}{*}{\rotatebox[origin=c]{90}{Success pct.}} & \multicolumn{1}{c|}{RRT} &  65.0\% & 90.0\% \\
 & \multicolumn{1}{c|}{AO-RRT} & 5.0\% & 55.0\% \\
 & \multicolumn{1}{c|}{Anytime-RRT} &  80.0\% & 95.0\%  \\
 & \multicolumn{1}{c|}{SST} &  85.0\% & 80.0\%  \\
 & \multicolumn{1}{c|}{SST*} &  55.0\% & 60.0\%  \\
 & \multicolumn{1}{c|}{AO-EST} &  10.0\% & N/A  \\ \bottomrule
\end{tabular}

\end{table*}

\section{Experimental Results}

We extensively validate L4KDE for tree-expansion using multiple dynamics models, the model dynamics examined include:
(1) Dubins car dynamics with condition-able turn rate; (2) Double pendulum dynamics; and (3) Gooseneck trailer dynamics.

The \textbf{Dubins car} dynamics are given by $\bb{x} = [x,y,\theta]$ and $\bb{u}=[u_s, u_\phi]$, where $\dot{x}=u_s \cos{\theta}$, $\dot{y}=u_s \sin{\theta}$, $\dot{\theta}=u_s/L_\text{axles} \tan{u_\phi}$, and $L_\text{axles}$ denotes the distance between the front and rear axles.
The \textbf{Double Pendulum} dynamics are given by $\bb{x} = [\theta_1,\omega_1,\theta_2,\omega_2]$ and $\bb{u}=[u_{\alpha_1}, u_{\alpha_2}]$, where $\dot{\theta_i}=\omega_i$, $\dot{\omega_i}=u_{\alpha_i} / (mL_i^2) - g \sin{\theta_i} / L_i$, and $L_i$ denotes the length of the $i^\text{th}$ arm.
The \textbf{Gooseneck Trailer} dynamics are given by $\bb{x} = [x,y,\theta_1,\theta_2]$ and $\bb{u}=[u_s, u_\phi]$, where $\dot{x}=u_s \cos{\theta}$, $\dot{y}=u_s \sin{\theta}$, $\dot{\theta_1}=u_s/L_1 \tan{u_\phi}$, $\dot{\theta_2}=u_s/L_2 \sin{(\theta_1 - \theta_2)}$, and $L_1,L_2$ denote the axles distance.

The L4KDE tree-expansion method can be used within a suite of widely-used tree-base kinodynamic motion planners, these include:
(1) RRT~\citep{lavalle2001randomized};
(2)Anytime RRT~\citep{ferguson2006anytime};
(3) SST~\citep{li2016asymptotically};
(4) SST*~\citep{li2016asymptotically}; and
(5) AO-RRT~\citep{hauser2016asymptotically}.
We additionally compared our planner with
(6) AO-EST~\citep{hauser2016asymptotically} which is incompatible with L4KDE because its tree-expansion mechanism measures the space's expansiveness.

All experiments are repeated 20 times, with a maximum run-time of 30 seconds for the Dubins vehicle, 3 minutes for the double pendulum and 5 minutes for the gooseneck trailer dynamics.
All our model use the same neural network architecture: state space input dimension, followed with 5 fully-connected layers each with ReLU activation functions in-between, and with a Sigmoid as the output activation function.
During training, we scale the target cost to cover the full range of the Sigmoid layer's output.
The metrics used in our experiments are: (1) \emph{time-to-solution}: the time, in seconds, taken to find a feasible solution; (2) \emph{success percentage}: the percentage of runs where feasible solutions are found.

\subsection{Dubins Car: Conditioning on Constraint Limit Values}

We wish to investigate whether L4KDE can generalise over an entire class of dynamics models, and predict the transition cost of Dubins car models with different kinematic constraints. We consider Dubins car models with different turn rate constraints, where $|{u_\phi}| \le \arctan{(L_\text{axles} \cdot \Omega)}$. The bounds on turn rate $\frac{\mathrm{d} \theta}{\mathrm{d} s}$ are denoted as $\Omega$. We collect transition cost data from Dubins car models where the turning rate limits, $\Omega$, are $15$ equi-distance values between $0.5$ and $\pi$. We then run L4KDE within the tree-based motion planners, where $\Omega=\{0.5,1.38,2.26,\pi\}$. An illustration of the test environment, along with an example solution (in red) and the tree (in grey), is given in  \cref{fig:dubins-car}. The results of using L4KDE, as compared to using Euclidean distance to select nearest nodes, within tree-based approaches is tabulated in \cref{table:turn-rate}, and the shortest time-to-solution plan for each $\Omega$ is given in bold. We observe that for different turn rate limits and over different motion planning methods, using L4KDE for tree expansion gives consistently shorter time-to-solution values and higher success percentages. The success percentages of each tree-based motion planner for each $\Omega$, over different run-times, are shown in \cref{fig:result-dubins}. We observe that the L4KDE variants of motion planning algorithms have a higher success rate at almost any given time than the standard variants.

\subsection{Gooseneck Trailer and Double Pendulum: L4KDE for Highly Challenging Systems}
We validate that L4KDE improves the quality of the tree-expansion in the challenging gooseneck trailer and double pendulum tasks. An outline of the environmental obstacle setup is shown in \cref{fig:dubins-trailer} and \cref{fig:2JP-start}. The time-to-solution and success percentage results are tabulated in \cref{table-hard-problems}. We see that for each of the different motion planning algorithms, using L4KDE significantly improves performance, often reducing the time-to-solution by more than half. The percentage of success over time for each method is illustrated in \cref{fig:result-other-env}. We observe that L4KDE variants of the suite of motion planning algorithms have a consistently higher success percentage over almost all points in time, for both challenging problem setups.

\begin{figure}[t]
    \centering%
    \includegraphics[width=\linewidth]{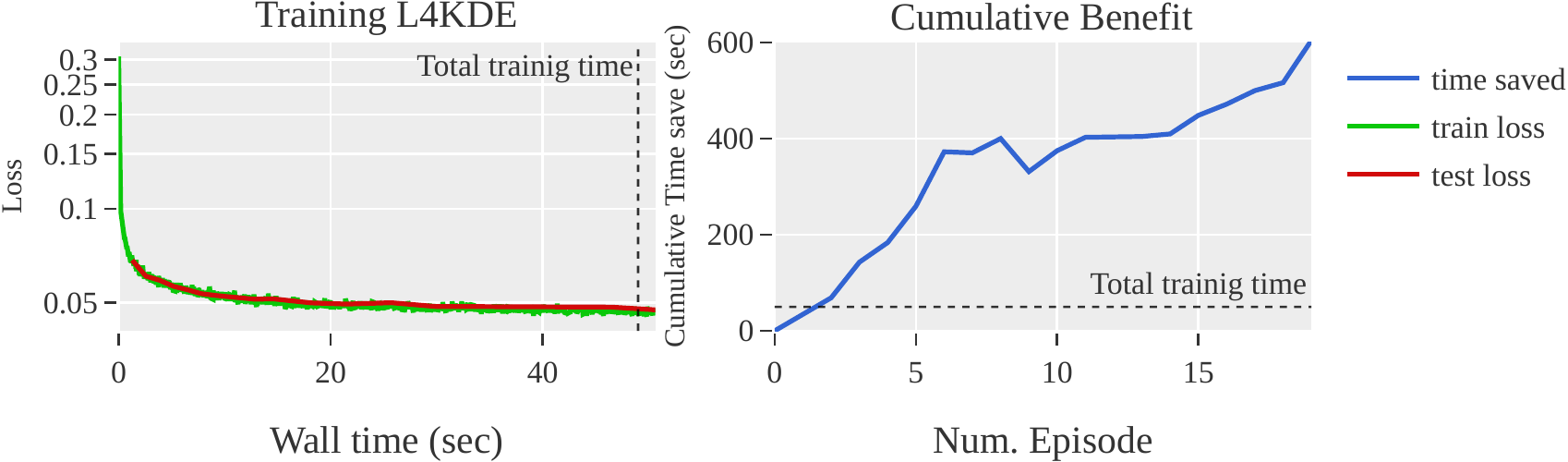}
    \caption{%
        The cumulative benefit of using L4KDE to accelerate planning after passively observing the system dynamics's transitions. Left plot shows the training loss against time in the \emph{double pendulum problem}. Right plot illustrates the corresponding time saved (cumulative differences of time-to-solution, $y_k = \sum_i^k (t_{\text{standard},i} - t_{\text{L4KDE},i})$) when using L4KDE for the tree expansion.
        \label{fig:l4kde-benefit}
    }
\end{figure}

\subsection{Offline Training time vs Cumulative Performance Improvement}
Training of the predictive model $g_{\theta}$ can be accomplished offline and as such is not time-critical to online planning performance. As such we can characterize training as a ``sunk cost'' before the online motion planning. We wish to contextualise the benefit of L4KDE with respect to the time that it took to fully train our predictive model.
In~\cref{fig:l4kde-benefit}, we record our model's training and testing loss against wall time for the double pendulum problem.
We then run multiple planning episodes using AO-RRT with and without L4KDE on the double pendulum problem setup, and record the cumulative time-to-solution differences between the two.
All procedures are performed on the same hardware.
We found that training our model can fully converge within one minute. This is the one-time cost of training.
Our sunk cost (in wall-time) can then be recovered within 1-2 episodes, and thereafter, the cumulative benefit of using L4KDE grows with every additional planning episode.
This illustrates the simple yet powerful cumulative benefit of utilising our learning-based method in kinodynamic problems, and the sunk-cost of training the predictive model can be recovered after a fairly small number of motion planning episodes.

\section{Conclusion}

We introduce \textbf{L}earning for \textbf{K}ino\textbf{D}ynamic Tree \textbf{E}xpansion (L4KDE) method for kinodynamic tree-based planning. Existing kinodynamic tree-based methods use Euclidean distance as a proxy for transition cost, when selecting the nearest nodes on the tree. Using Euclidean distance for node selection is fast, but often leads to suboptimal nodes being selected. We propose to instead train a neural network to efficiently predict the transition cost online. Our model is able to condition on parameters and limits of the dynamics model, enabling us to predict the costs over an entire class of model, rather than a single model. We demonstrate that utilising L4KDE for tree-expansion provides significant performance improvements in terms of success rate for a given time budget and time to solution within a suite state-of-the-art asymptotically optimal, tree-based motion planning methods. We additionally show that the upfront cost of training our model quickly pays for itself after only a few online motion planning episodes in terms of overall compute time saved.

\bibliography{ref.bib}

\end{document}